\documentclass{article}
\usepackage{graphicx}

\begin{document}
\title{Core Conflictual Relationship: Text Mining to Discover What and When}
\author{Fionn Murtagh and Giuseppe Iurato}
\maketitle

\begin{abstract}
Following detailed presentation of the Core Conflictual Relationship Theme
(CCRT), there is the objective of relevant methods for what has been described as 
verbalization and visualization of data. Such is also termed data mining and 
text mining, and knowledge discovery in data.  The Correspondence Analysis 
methodology, also termed Geometric Data Analysis, is shown in a case study 
to be comprehensive and revealing.  Computational efficiency depends on how 
the analysis process is structured. For both illustrative and revealing aspects
of the case study here, relatively extensive dream reports are used.  This Geometric
Data Analysis confirms the validity of CCRT method.  
\end{abstract}


\section{Introduction}

The Core Conflictual Relationship Theme is quite essential in psychotherapy. 
There are individual's relationships with his or her parents, and with other
established personages, all of these relationships can and will be carried over 
into other relationships and also into other behavioural patterns.  In Aylward (2012),
terrible social violence is described and related to the perpetrator's youth and 
other relations and behaviours arising from that.  Murtagh and Iurato (2016) also 
relate behaviours, and so on, with therapy.  The related themes involved in these 
interpersonal relationships have been ascertained to be almost universal and commonly 
shared among all humans, and this is conferring objectivity to the attempts to identify 
them for instance either qualitatively via CCRT method and quantitatively via Geometric 
Data Analysis, as just we have done in this work. A similar methodology has been already 
performed in Murtagh and Iurato (2017).

A lot of what we are dealing with in this work is the quantitative (by Geometric 
Data Analysis) as well as 
qualitative (by CCRT approach) analytical methods.  In the next section 
there is a description of 
CCRT and how it may quantify by Geometric Data Analysis various aspects of 
psychotherapy processes and procedures. 
In the section entitled ``Implementation and Objectives'', 
there is outlining of the textual, hence descriptive, 
data used in this work.  The analysis work starts in the section 
entitled ``Analytical focus on selected names'', 
with dream reports of views and relationships with parents, children, friends, brothers, 
and a person who had been a husband and after divorce, he died. 
That is general analytical processing.  In the sections entitled ``Associating with 
a named individual'', and ``Study of mother'', 
there is a further mapping out of emotional relationships, described in the dream reports.  

\section{Core Conflictual Relationship Theme, and Data Mining Analysis}
\label{sect2}

The CCRT method, i.e., the Core Conflictual Relationship Theme method, based on 
the text analysis of the single narratives coming from an analytical setting, 
enables a transposition of the concept of {\em transference}, that is, one 
of the chief notions of Freudian psychoanalysis, from the level of understanding, 
i.e., the epistemological key of human sciences methodology, to the level of 
explaining, i.e., the epistemological key of natural sciences. This 
allows to get the epistemological transformation of a 
concept of humanities into a concept of natural sciences, as it becomes 
liable to be measurable and assessable according to the usual methodology of 
the latter disciplines. Expressed alternatively, we relate both qualitative and 
quantitative analysis by integrating both Correspondence Analysis and CCRT method.

The CCRT method is based on the assumption that 
transference pattern is structurally rooted in every human being, moulded in 
infancy either by primary identifications with caregivers (usually, parents) 
and by innate factors, a pattern which is generally deemed as almost persistent 
as such in time, along the whole life's course of every human being, from 
which alternativeness (or otherness), i.e.\ the primary sense of the other, which 
goes beyond one's own individuality and self-love, arises, as well as almost the entire 
personality (Luborsky and Crits-Christoph 1990, Chapters 1, 18). 

The CCRT method basically springs out from the observation of the recurrence of 
three main {\em leitmotivs} or themes during each analytical setting: what the 
patient wishes from others, how other persons accordingly react, and how the 
patient replies to these latter reactions. Therefore, Lester Luborsky (1920--2009), 
in the 1970s, identified the three main categories corresponding to these three 
recurrent themes, that is to say: the intentions, needs and desires toward another 
person, the corresponding responses of the other person, the consequent answers of 
the own Self. The final CCRT is given by the suitable combination of the most 
pervasive components present (and identified by means of Geometric Data Analysis) 
in each of these categories, retrieved from the 
various stories stemming from analytical setting. 

The attention directed at  
these three main recurrent themes, is then supported by the previous psychoanalytic 
researches and studies which have shown that, in almost every human being, the 
fantasies are regrouped around some chief basic desires and the related 
intrapsychic conflicts. These fantasies take place early, since childhood, with 
their thematics remaining the same for the whole life cycle, at most changing only 
their scene or the personages therein involved, just like in the CCRT (Luborsky, 1984, 
Luborsky and Crits-Christoph 1990, Chapter 1). 

Therefore, all this gives rise to the assumption that there exist deep and 
rooted unconscious fantasies which are almost universal and commonly shared by 
all human beings, outcomes of ancestral models of relationships deeply ingrained 
into human psyche which become active when one encounters the other. See above all,
Laplanche (1985), by  Jean Laplanche (1924--2012, associate of Jacques Jacan), who 
in an interview 
(Laplanche, 1999) states this: ``primary symbolic identification, is 
not `I identify myself' but an identification by the other. The other identifies me.''

On the other 
hand, in Freud's earlier work, from 1912, recurrent thematic units were identified 
in the transference settings, made by dominant needs into reciprocal relationships 
established during childhood, along which these are taking, or not, a conflictual 
nature, and they  are destined to be repeated again, during the next, whole life 
course, characterizing the related psychic development. 

Afterwards, Luborsky considers a comprehensive set of other psychological studies 
which confirm, by analogy, his model, which historically sprung out from various, 
previous attempts to measure transference. From these, Luborsky and co-workers noted 
the occurrence of certain pervasive or recurrent themes during the analytical 
setting and related therefore with interactions between Self/Other. For these 
reasons, the method was named Core Conflictual Relationship Theme, as it has to 
do with central (i.e., Core) recurrent themes emerging from the various 
psychodynamic conflicts (i.e., Conflictual) established by the possible relations
 (i.e., Relationship) between the Self (patient in psychotherapy) and the Others 
(analyst, as well as other persons). From CCRT method on, an operative evaluation 
of transference phenomena has been then pursued (Luborsky and Crits-Christoph 1990, 
Chapter 1). 

From the operational stance, the CCRT method starts with the identification of 
narrative units, said to be {\em relational episodes} (expressed as, RE), in which 
the patient is particularly involved in relationships with others (analyst, 
parents, friends, etc.) in a typical and primary manner, until putting into 
action the episode itself during the psychotherapy being undertaken. The RE 
should be complete, in the sense that it should be described in a complete manner,
 above all in respect to the various situations related to the relations involved 
in such an RE. 

Once an RE has been identified, the next step is to identify, in 
a given RE, the various {\em thought units} composing such an RE, that is, 
the principal propositions present therein, hence the analyst proceeds with 
the identification of the major components of an RE, that is: wishes, desires, needs and 
intentions  (W); responses of the others (RO); responses of the Self (RS), oneself. 
Afterwards, one proceeds to identify the various (implicit or explicit) meanings of 
the thought units, as, for example, the possible affective states involved there, 
as well as to classify all the possible responses of either the others (RO) and 
oneself (RS), classifying them as positive or negative, attended (i.e., not 
realized) or actuated (i.e., realized). Each of these items is then classified 
with a related score assigned by the examiner (supporting the therapist) who 
then should identify too the 
related recurrence (Luborsky and Crits-Christoph 1990, Chapter 2).

What is important in the compilation of CCRT, is the identification of the 
components W, RO and RS in each RE, to count these, to classify these as 
`positive or negative' for RO and RS, as well as to classify RO as 
`expressed in action or simply attended'. Hence, the examiner goes on to 
assign, to each component, a score in dependence on the frequency of its 
occurrence, in such a manner that, after this, one is able to describe what 
type of components (usually, of W type, but not limited only to this) are 
more frequent than others in the whole set of RE, with the principal aim to 
identify the final CCRT around which the main psychodynamic conflicts take 
place. Therefore, the final CCRT is provided by the most frequent themes of 
the type W, RO and RS as detected into the whole series of RE. As regards, 
then, W type themes, these should be previously determined in dependence on 
their degree of inference, which may be explicit (We) or more or less implicit 
(Wi), the latter being more frequent than the former, hence much more important 
from a psychoanalytic standpoint for the latent meanings brought by them. For 
this, the examiner should identify as many as possible W type items, in that 
they are (as unsatisfied) the main centres around which revolve psychic 
conflicts. Much easier is the identification of RO and RS types, as they are 
usually expressed directly and are consequence of desires in that they are 
closely related to the satisfaction or not of desires and needs, or rather 
to their alleged satisfaction  (whence, their classification as positive or 
negative). Finally, the themes of the components W, RO and RS are also 
classified in dependence on their intensity (Luborsky and Crits-Christoph 1990, 
Chapter 2).

The pivotal linking point between CCRT method and Freudian transference is just 
the parallel that should hold respectively between, on the one hand, the 
conflictual pair desires/needs/intentions (W) versus responses (RO/RS) and, on 
the other hand, the conflictual (Freudian) pair Es' pushes (impulses/drives/desires) 
versus Ego's responses (as outcomes of the defence's mechanisms), and this congruence 
should need a validation on the basis of the assessment of CCRT method by therapeutic 
outcomes coming just from the clinical applications of CCRT method, which has shown 
that the components W, RO/RS have a high frequency of association and that such 
association has just a conflictual nature. So, having seen the high degree of objectivity 
of CCRT method, from this last result should follow that also Freudian transference 
should have or gain the same degree of objectivity if one were able to show the 
subsistence of the above crucial congruence between Freudian model of transference 
and CCRT method. Maybe, data mining analysis, such as the most coherent geometric 
data analysis and semantic mapping provided by Correspondence Analysis might bring aid 
or shed light to solve this last question or, at least, make more explicit such a 
fundamental link between Freudian transference model and CCRT method (Luborsky and 
Crits-Christoph 1990, Chapter 18).

\section{Our Implementation and Objectives}
\label{sect3}

On account of the private or confidential nature of psychotherapy sessions, here
we use the following data to explore the analytical processing and the potential 
for obtaining outcomes that will be relevant and important for the objectives 
described in the previous section (intentions, needs and desires towards another 
person, with the consequences and related responses).  

Used are much more data comprising dream reports where previous analysis carried out 
was in regard to individual relationships, Murtagh (2014a, 2014b).  This data from 
Barbara (mostly written as Barb) Sanders is from DreamBank (2004), see also 
Domhoff (2003, 2006).  Related 
analysis work on this textual expressing of dreams by this individual, Barb
Sanders, is extensively covered in much of Domhoff (2003) and, in particular, in 
chapter 5.  In chapter 4 (page 99), the frequency of occurrence of naming pet 
animals leads to Barb Sanders being a ``cat lover'' rather than a ``dog lover''. 

A point about our Correspondence Analysis methodology is that semantic similarity 
or identity is very supported through such terms being semantically mapped close to 
each other, or even potentially, superimposed in the factor space mapping. 
In Domhoff (2003, chapter 4), there is the noting that a cat lover may be using 
these terms: 
``cat'', ``kitten'', ``kitty'', ``kittie'', ``feline''; and a dog lover uses 
these terms: ``dog'', ``doggy'', ``doggie'', ``puppy'', ``puppies'', ``canine''.  
Opposite to this semantic commonality is supporting disambiguation, i.e.\ that 
identical or very similar spelling could be the case for quite distinct words, so their 
semantic mapping must have them distinct in the factor space.  In Murtagh (2015) it
is shown how misspellings are likely to be closely related in their locations 
in the factor space, and also singulars and plurals of words.  The reason why 
lemmatization is not applied here (as it often is in the textual analytical 
processing) is that it may be the case that variations in such 
grammar can be revealing in its distinctiveness.  

In Domhoff (2006) there is the following description. This quotation is from it: 
``Dreams are dramatizations, or enactments, if you will, or our thoughts''. 
Another quotation: ``She had several boyfriends after her divorce and never remarried.''
Born in the 1940s, she ``did not start a dream journal until a few years after her 
divorce''.  There is description of the great importance for her of her mother, 
the importance also of her father, and how her ``middle daughter'', at 4.5 years of 
age, reacted to her divorce; then, ``By contrast, Sanders dreams only half as often as
her oldest and youngest daughters, who adjusted to the divorce better''; and there is 
description of the brother closest in age to her, and friends.  Close friends include
Ginny and Lucy.  Discussion includes brother, Dwight, friend, Darryl. 

In Domhoff and Schneider (2008), reference is made to the 22,000 dream reports 
available at DreamBank.net, of which 16,000 are in English.  Characters in dreams
are noted as being described by power laws (i.e., exponential distributions, and 
what is referred to as Zipf's law in information retrieval). Figure 1 relates to 
Barb Sanders. In analysis, used here are dream reports with the highest rank, in 
the dream contents, relating to the 
mother, then the father, then the oldest daughter, and next the following: middle 
daughter, 
youngest daughter, favourite brother, friend Ginny and friend Lucy.  There is further
discussion of religion, and sexual activity in dreams.  A great deal more discussion 
is in regard to the substance and consistency of dreams.  Also discussed are 
expressions of emotion from: ``anger, apprehension, sadness, confusion, and happiness''.
Reference is also made (section 3.7) to the appearance of these expressions in dream 
reports: 
``my dad'' and ``my mom''.  For Barb Sanders there is this (section 3.8): ``Sander's 
rather 
perfunctory conversion to the Episcopal church when she married her husband, many 
years before she began to write down her dreams''. It states (in section 4) that, for 
quantitative analysis, more than 125 dream reports are to be required. 

Our objective is to map out what can be of central analytical importance using 
the Barbara Sanders dream reports.  Note that the name Barb is more the case in the 
references (and in some of the dream texts, there is the self-reference by this 
individual that just uses the letter, B). From DreamBank (2004), dream reports were 
obtained.  In all there are 3116 dream reports available, from the years 1960 to 
1997.  Using a listing from DreamBank (2004) entitled ``The ``cast of characters'' in 
the Barb Sanders dream series'', that listed 125 names, each with their gender and 
their ``Relationship to Barb Sanders'', the following names were selected here: Darryl, 
boyfriend; Derek, male friend; Dovre, daughter (oldest); Dwight, brother; Lucy, female friend; 
my father, father; my mother, mother; Paulina, daughter (youngest); Ellie, daughter (middle); 
Ginny, female friend (married to Ernie); Howard, ex-husband; Jake, brother.  Our motivation 
for this selection was to have mother and father, ex-husband, all daughters (she 
had no sons); her brothers (her one sister was not included here), two friends (quite
a few others not included), and a boyfriend (and eight others not included here).  
The ex-husband, Howard, died in 1997.  

Taken for this analysis from the 1106 dream reports relating to the above listed names,
were 421 of these.  
The 421 dream reports here are from index number 4, from 2 December 1960 to index number 
1264, from 17 February 1989.   
Each dream report varies from a few words to about 900 words.  In total, the 421 dream 
reports, in succession between 2 December 1960 to 17 February 1989 have 3789 words.

Our aim here is to have a general approach for this analysis, and this can always
be complemented with specific and derived procedures, possibly using statistical 
modelling or machine learning.  In the referenced Barb Sanders, ``Dreams and Waking 
Life: Interview Information ...'', much information is included about her personal 
life and relations.  

\section{Analytical Focus on Selected Names}
\label{sect4}

The data consists of frequency of occurrence values, encompassing presence or 
absence where the latter have frequency of occurrence values of 1 or 0.  For the 
421 dream reports, and the initial word corpus of 6376 words, we require each word 
to be used at least five times.  That is so as to exclude exceptionally used words
from consideration, and rather to have a requirement for some degree of commonality 
of word use.  These result in a word corpus of 1568 words, for the 421 dream reports. 
The crossing (i.e.\ a frequency of occurrence matrix, including especially presence 
and absence values), of 421 dream reports by 1568 words, this has 37344 non-zero 
(i.e.\ non-absent) values, which is 5.66\% of all values.  

Some of the dream reports thereby become empty, sixteen of them, so therefore our 
analysis on dream reports of sufficient length is to be on 405 dream reports, with 
frequencies of occurrence for the 1568 word set.  From this word set, the selected
names are to be main focus: mother, father, Ellie, Howard, Dwight, Paulina, Ginny,
Dovre, Darryl, Lucy, Jake. 

\begin{figure}
\includegraphics[width=12cm]{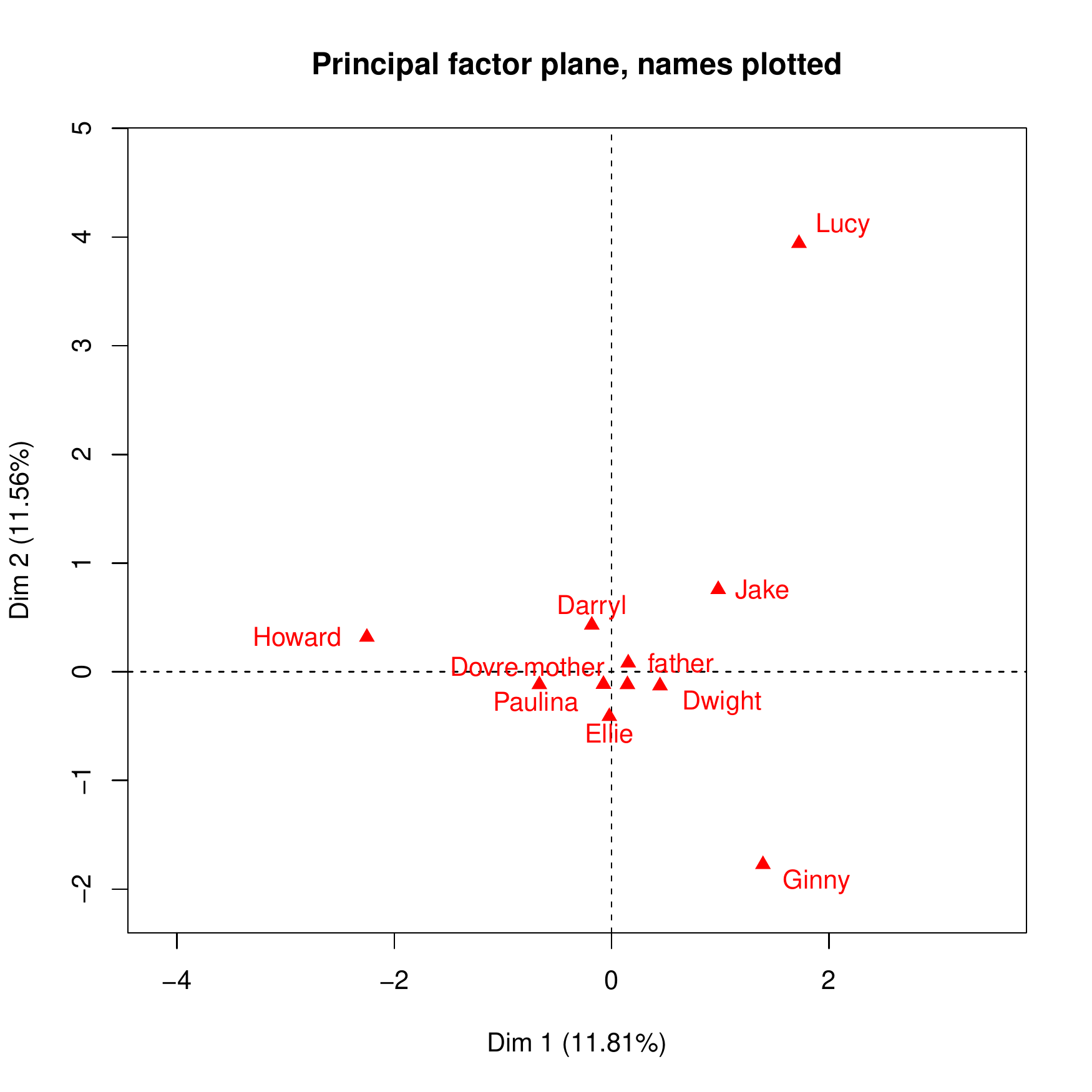}
\caption{Factors 1 and 2, displaying just the names here.  These are the active 
variables in the analysis.}
\label{fig1}
\end{figure}

\begin{figure}
\includegraphics[width=12cm]{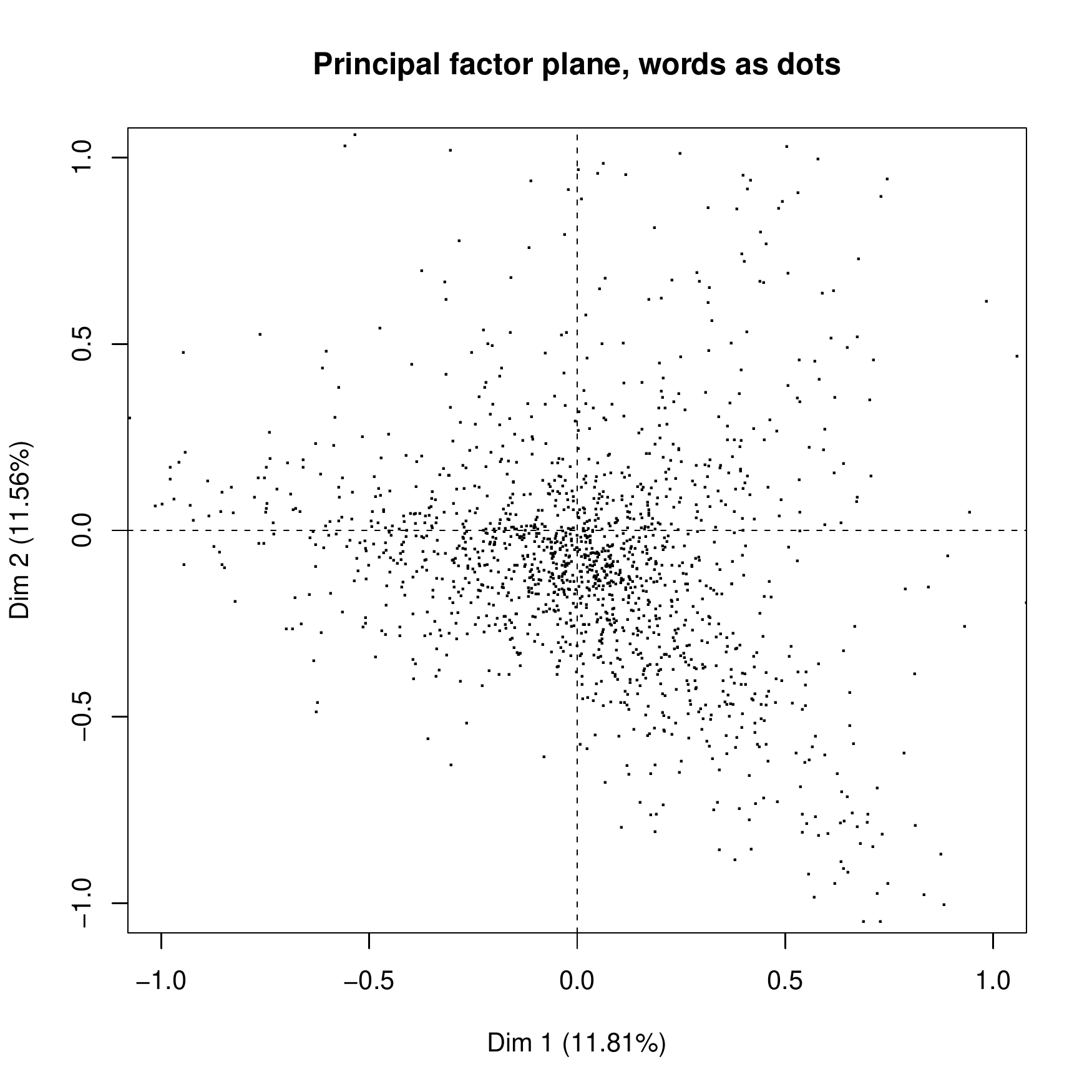}
\caption{Displaying the 1557 words.  Relative to Figure \ref{fig1} and 
the scaling displayed for factors 1 and 2, i.e.\ the horizontal and vertical
axes, here there is more concentration.}
\label{fig2}
\end{figure}

\begin{figure}
\includegraphics[width=12cm]{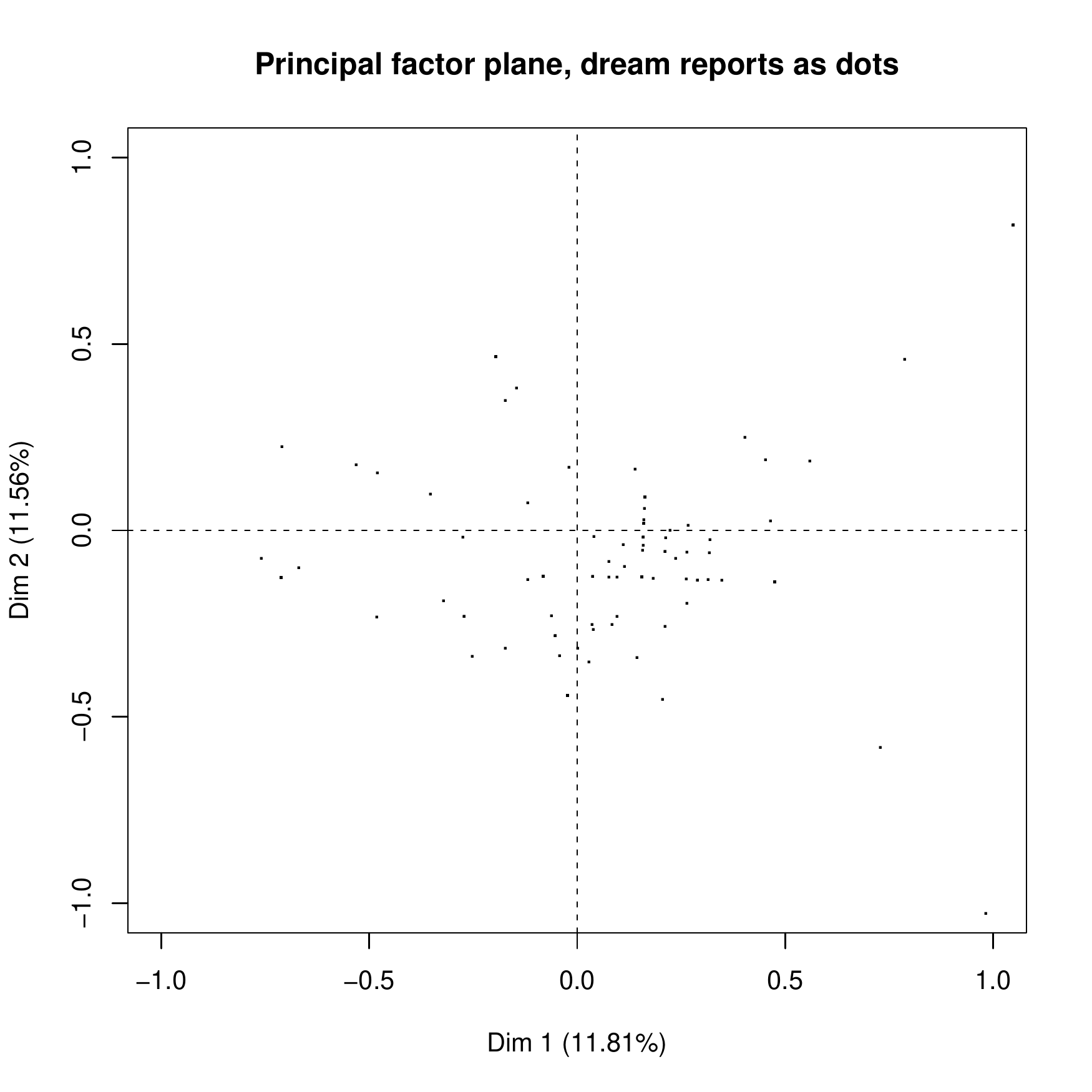}
\caption{Displaying the 405 dream reports.}
\label{fig3}
\end{figure}

For the orientation of the analysis, or what we might term the focus of the analysis,
these names are selected from the set of words. They constitute the active variables
(i.e.\ selected words), in the semantic mapping.   
Figure \ref{fig1} displays the principal factor plane.  The eigenvalues 
that express inertia of the factors, in percentage terms are: 11.8, 11.6, 11.3, 11.0,
10.7, 9.8, 9.6, 8.6, 8.1 and 7.7.  In order to look further at the words expressing 
the dream content, just the dots represent the word locations in the principal factor
plane, displayed in Figure \ref{fig2}.  The changes over time are also to be looked at
further, and these are simply displayed in Figure \ref{fig3}. 

Here, as follows, are the contributions by the selected names to inertia of the five factors.  
It is seen that factor 1 is most essentially relating to ex-husband Howard, and friends
Ginny and Lucy.  Factor 2 is most essentially relating to Ginny and Lucy.  Factor 3 
is most essentially relating to friend Ginny, daughter Ellie and ex-husband Howard. 
Factor 4 is essentially relating to daughters Dovre and Ellie, boyfriend Darryl and 
father.  Finally here, as follows, factor 5 is essentially relating to Darryl and father. 

Contributions of the selected names to the five factors:

\begin{verbatim}
        Dim 1 Dim 2 Dim 3 Dim 4 Dim 5
mother    0.4   0.3   1.7   7.0   2.7
father    0.5   0.1   1.8  16.2  23.3
Ellie     0.0   2.3  19.9  15.3   0.1
Howard   61.7   1.3  21.6   0.0   0.3
Dwight    2.3   0.2   0.1   1.7   0.0
Paulina   0.1   0.1   9.7   3.6   2.3
Ginny    16.7  27.6  37.5   2.9   1.5
Dovre     2.8   0.1   1.0  27.4   1.1
Darryl    0.2   0.9   0.5  19.3  68.2
Lucy     12.2  65.0   4.9   6.6   0.1
Jake      3.3   2.0   1.2   0.0   0.2
\end{verbatim}

Here are the coordinates on the five factors of the selected names:

\begin{verbatim}
        Dim 1 Dim 2 Dim 3 Dim 4 Dim 5
mother    0.1  -0.1  -0.3  -0.6  -0.3
father    0.2   0.1  -0.3  -0.9  -1.0
Ellie     0.0  -0.4  -1.2   1.0   0.1
Howard   -2.3   0.3   1.3   0.1  -0.2
Dwight    0.4  -0.1   0.1  -0.4   0.0
Paulina  -0.1  -0.1  -1.0   0.6   0.5
Ginny     1.4  -1.8   2.0   0.6   0.4
Dovre    -0.7  -0.1  -0.4   2.0   0.4
Darryl   -0.2   0.4  -0.3  -1.9   3.5
Lucy      1.7   3.9   1.1   1.2   0.1
Jake      1.0   0.8   0.6   0.0  -0.2
\end{verbatim}

\begin{figure}
\includegraphics[width=12cm]{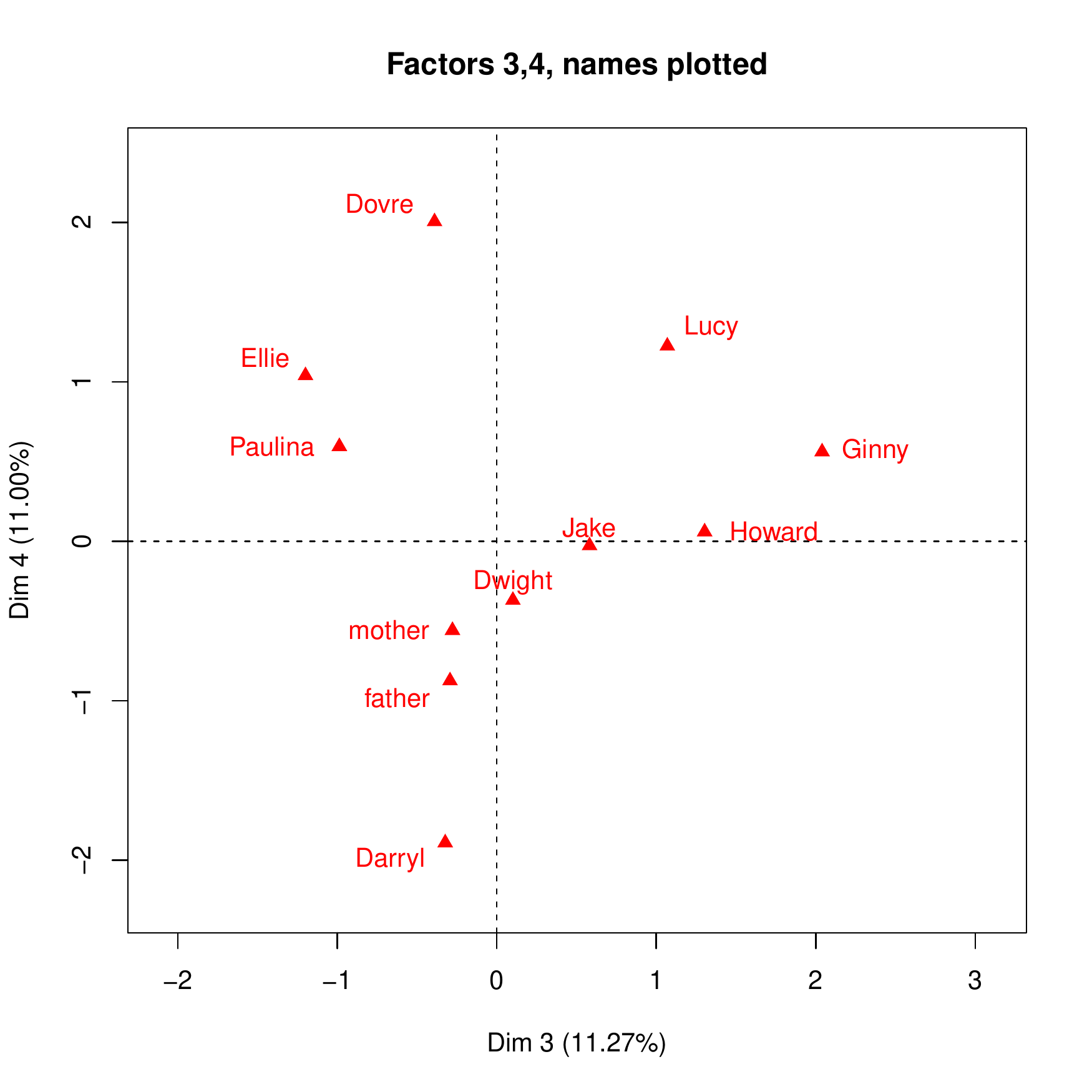}
\caption{The names in the factor plane of factors 3 and 4.}
\label{fig4}
\end{figure}

Figure \ref{fig4} shows the next factorial plane, the plane of factors,
or principal component axes, 3 and 4. 

Let us look now at what is most dominant for all of the factors.  Since 
the active analysis is on eleven names, crossed in terms of frequency of 
occurrence (with 0 frequency of occurrence implying absence of this word
in the dream report) with 405 dream reports.  The supplementary variables
are the 1557 other words here.  We can term supplementary variables, like
here, as contextual variables, i.e.\ providing the context for the names
(the set of eleven names that are the active variables here).  Active
variables are, effectively, the primary focus of the analysis.  Therefore
the active, focussed analysis is for 405 dream reports crossed by 11 names; 
and the supplementary mapping is for the 405 dream reports crossed by the 
1557 other word corpus members. 

Here are the contributions by all eleven names to the entire set of factors.  
 
\begin{verbatim}
        Dim 1 Dim 2 Dim 3 Dim 4 Dim 5 Dim 6 Dim 7 Dim 8 Dim 9 Dim 10
mother    0.4   0.3   1.7   7.0   2.7   0.0   0.0   0.1  18.6   50.9
father    0.5   0.1   1.8  16.2  23.3   0.0  16.7   0.6  15.2    8.5
Ellie     0.0   2.3  19.9  15.3   0.1  13.5   7.3  28.7   0.2    1.0
Howard   61.7   1.3  21.6   0.0   0.3   2.0   2.3   0.0   0.1    0.1
Dwight    2.3   0.2   0.1   1.7   0.0   6.8   8.5   1.6  32.7   36.2
Paulina   0.1   0.1   9.7   3.6   2.3   2.9   1.6  64.5   6.8    0.0
Ginny    16.7  27.6  37.5   2.9   1.5   3.3   1.5   0.0   1.4    0.1
Dovre     2.8   0.1   1.0  27.4   1.1  34.8  26.4   0.2   0.6    0.1
Darryl    0.2   0.9   0.5  19.3  68.2   0.0   2.1   2.8   1.5    0.0
Lucy     12.2  65.0   4.9   6.6   0.1   5.7   1.2   0.0   0.7    0.0
Jake      3.3   2.0   1.2   0.0   0.2  31.0  32.3   1.5  22.4    3.0
\end{verbatim}

Briefly, we see that factor 1 is predominantly related to Howard; factor 
2 is predominantly related to Lucy; factor 5 is predominantly related to 
Darryl; factor 8 is predominantly related to Paulina; and factor 10 is 
predominantly related to mother. 

\begin{figure}
\includegraphics[width=12cm]{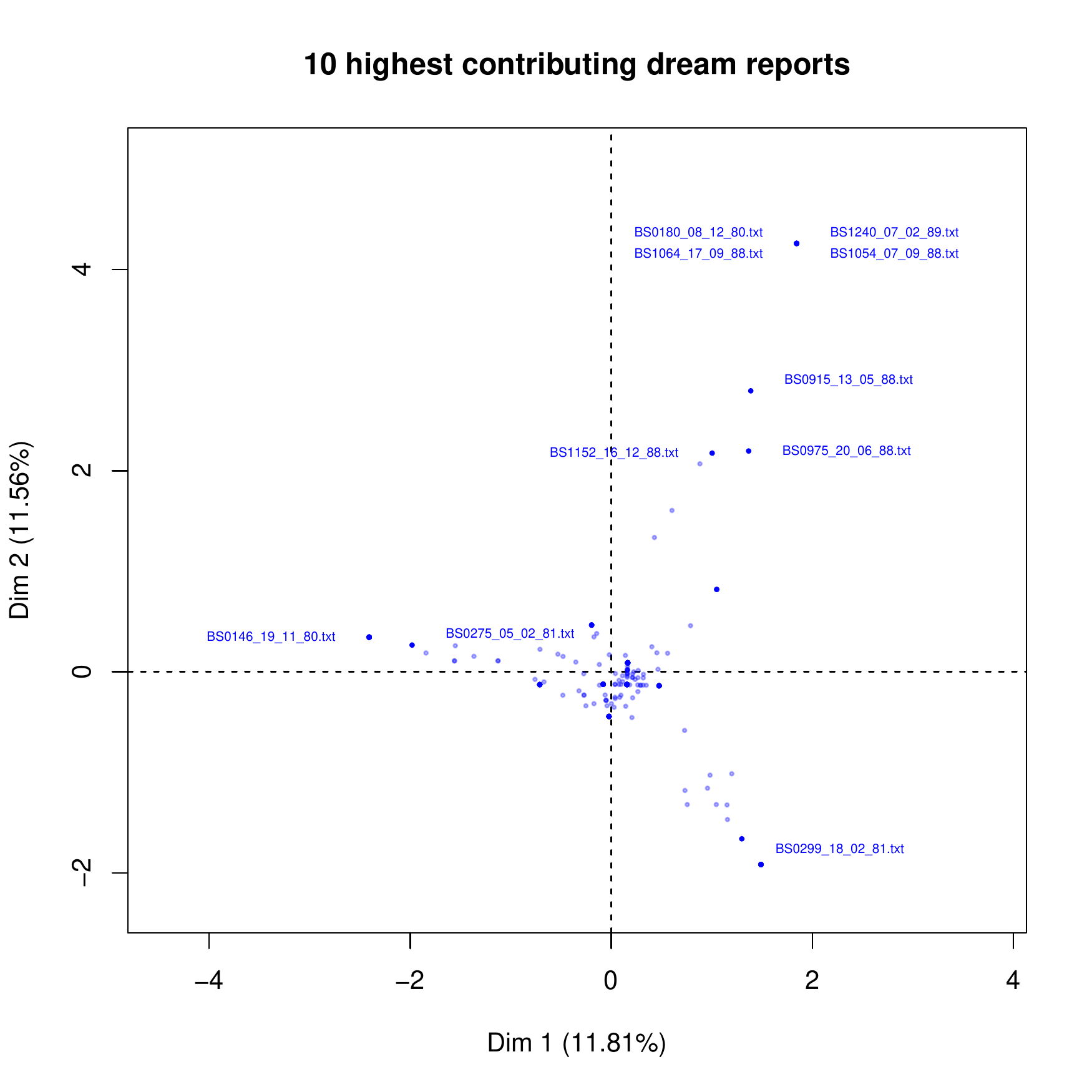}
\caption{The ten highest contributing (to the inertia of factors 1 and 2) 
dream reports in the factor 1, factor 2 space. These names have the letters
BS, then a sequence number, and then the data (with format: day, month, last two 
digits of the year).}
\label{fig5}
\end{figure}

\begin{figure}
\includegraphics[width=12cm]{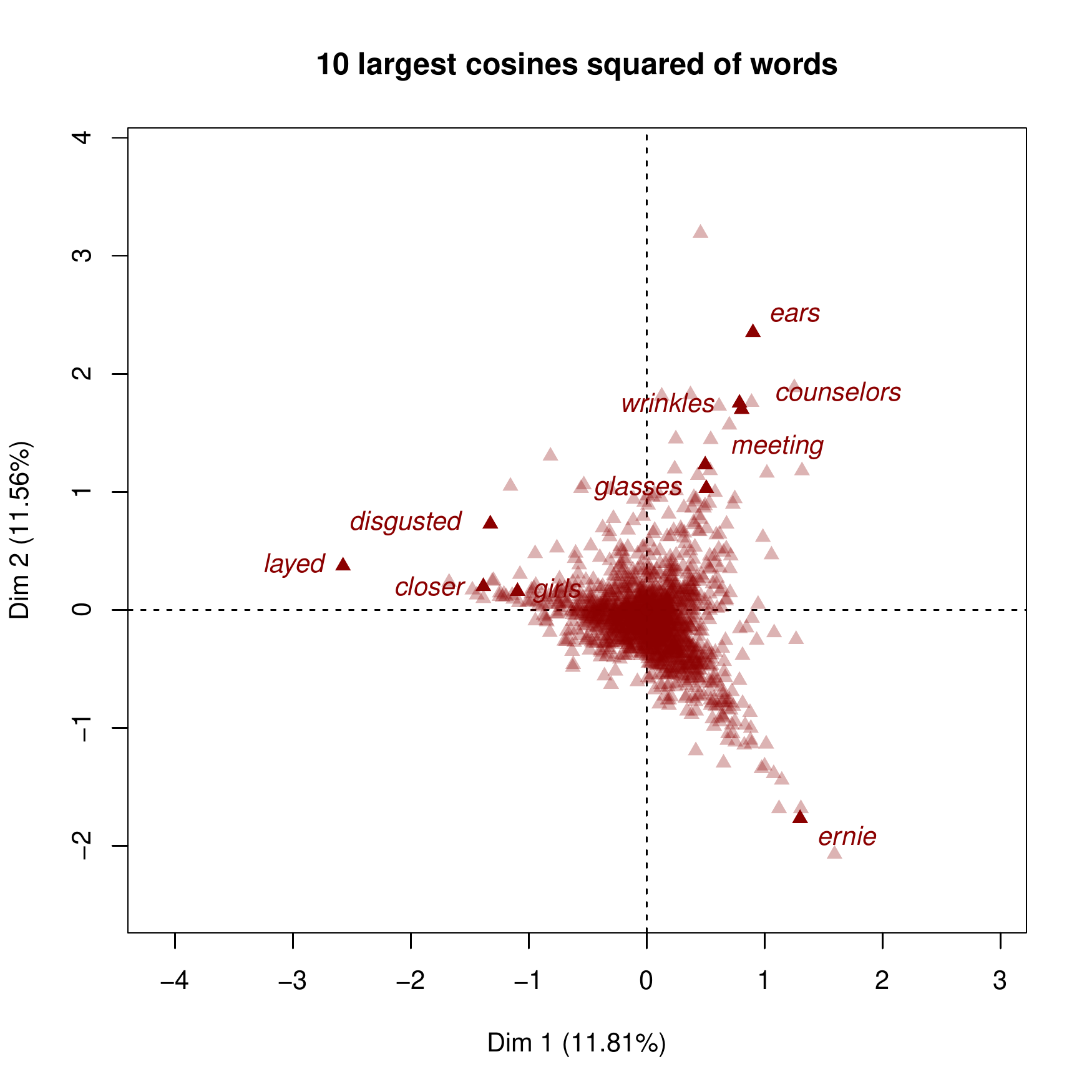}
\caption{The ten highest contributing words from the word corpus (in this analysis, 
taken as supplementary variables, relative to the active variables comprising the 
eleven names).}
\label{fig6}
\end{figure}

We can seek to find the association and relevant informative relationships, 
as displayed for Factors 1 and 2 in Figures \ref{fig5} and \ref{fig6}. Figure
\ref{fig5} displays this for the ten most contributing (to inertia of the
factors, i.e., the axes) dream reports.  Figure \ref{fig6} displays the ten 
words from the word corpus derived from the data, with the highest squared
cosines, and thereby what is in effect the correlations with the axes. 

\section{Associating with a Named Individual: Verbal Expression -- What and When}
\label{sect5}

While referring to verbal expression, this can and will be always put into text, 
through being digitally recorded or digitally stored. 
Verbal expressions here are single words, members of the word corpus that is derived from 
the set of texts that comprise the data under analysis.  As noted above, semantic 
proximity or near identity is well handled.  Semantics here is the entirety of 
relationships.  

We seek here to derive the most salient, the most informative and the most revealing
expressions relating to the selected set of individuals. Again it is noted that these 
eleven individual names are:  ``mother'',  ``father'',  ``Ellie'',  ``Howard'',  
``Dwight'',  ``Paulina'', ``Ginny'',   ``Dovre'',   ``Darryl'',  ``Lucy'',    ``Jake''.
The previous section entitled ``Our Implementation and Objectives'', explains these 
individuals.  The third onwards here are: daughter, 
ex-husband, brother, daughter, friend, daughter, boyfriend, friend, and brother. 

We select the ten semantically closest words to the name.  These are closest
to the name but clearly they can possibly be used also with other names. So for each
of the words, we seek the use of the word for that name; and list the dream record
dates when they are being used. That will allow us, firstly, to determine what are
the most associated words relating to dream reporting for that name.  Secondly, the
pattern of word use over time can be followed. To some extent, this may help to
know if such patterns in word usage can be associated with emotional expressions, or
the nature of the relationship with the individual of that name.

For the processing carried out, all words have their upper case letters set here 
to lower case.  The distances, between each of these words and the name, are shown 
in order to quantify the ranking of their semantic importance for the name.  Note
that such calculations are carried out in the full dimensional factor space. 
The dream report dates are to consider any variation over time.  
The distances, overall, follow a normal distribution.  (That was tested for the 
unique distances, i.e.\ for the dream reports, 81,810 distances; for the eleven names,
55 distances; and for the supplementary columns that comprised the word corpus, 
1,211,346 distances.  Also tested were all unique distances here, i.e.\ 
194,626,585 distances.  The mean distances, respectively, are, in two decimal places
of precision: 3.91, 4.67, 1.70 and 0.56. The Shapiro-Wilks test of normal distribution
gave a p-value in all cases of 2.2e-16.)    

\subsection{Name: mother}

The ten closest words in the semantic, full-dimensional factor space, with their 
distances, here rounded to two decimal places, are: 

\begin{verbatim}
doors sleepy stood nose nurse invisible clearly engagement king mid 
 0.54  0.43  0.48  0.49  0.53  0.53       0.46      0.53   0.53 0.53 
\end{verbatim}

The number of dream report texts with this name, from the 405 dream reports, is 108.
In fact, 24 of the dream reports contain one or more of the ten 
semantically closest words.

\subsection{Name: father}

\begin{verbatim}
paul counter shoe tom hardly josh harrison bricks salesman ship 
0.28   0.61  0.76 0.45 0.75  0.47   0.80    0.76    0.80   0.48 
\end{verbatim}

The number of dream report texts with this name, from the 405 dream reports, is 85.
In fact, 39 of the dream reports contain one or more of the ten 
semantically closest words.

\subsection{Name: Ellie, daughter}

Although ten closest words are at issue, here there is one distance repeated so 
this leads to eleven closest words at issue here. 

\begin{verbatim}
tyler hers kathleen jungle reluctantly beads cereal scarf snuck scream captain
0.61  0.64   1.04    0.71    1.18      0.81   1.21   1.08  1.21  0.88    0.67 
\end{verbatim}

The number of dream report texts with this name, from the 405 dream reports, is 55.
In fact, 23 of the dream reports contain one or more of the ten 
semantically closest words.

\subsection{Name: Howard, ex-husband}

\begin{verbatim}
closer trapped relieved disgusted wanting forward layed accidentally please bob
1.25    1.31    1.50     1.10      1.51    1.44    0.47   1.16        1.42  1.43
\end{verbatim}

The number of dream report texts with this name, from the 405 dream reports, is 61.
In fact, 24 of the dream reports contain one or more of the ten 
semantically closest words.

\subsection{Name: Dwight, brother}

\begin{verbatim}
brother teach beat teaching bartender monkey xmas plate surgery marry
 1.21   1.19  1.05  1.21     1.10      0.94  1.02 0.60   1.02   1.02
\end{verbatim}

The number of dream report texts with this name, from the 405 dream reports, is 53.
In fact, 39 of the dream reports contain one or more of the ten 
semantically closest words.

\subsection{Name: Paulina, youngest daughter}

\begin{verbatim}
fletcher mad listening manager journey  rope  catches cross branch button
 0.76   1.03   1.33     0.63    1.29    0.91   0.76   1.08   0.46   0.75
\end{verbatim}

The number of dream report texts with this name, from the 405 dream reports, is 49.
In fact, 23 of the dream reports contain one or more of the ten 
semantically closest words.

\subsection{Name: Ginny, friend, female}

\begin{verbatim}
ball disabled ernie raul repair deaf signs actors shore diane
1.54   1.51   0.37  0.90  1.13  0.98 1.48  0.57   1.25   0.67
\end{verbatim}

The number of dream report texts with this name, from the 405 dream reports, is 35.
In fact, 18 of the dream reports contain one or more of the ten 
semantically closest words.

\subsection{Name: Dovre, oldest daughter}

\begin{verbatim}
air gathering opened kittens rail spanish interest esther works chute
1.92   1.75    1.61    1.96  1.36  2.03    1.99     1.90  1.71  1.67
\end{verbatim} 

The number of dream report texts with this name, from the 405 dream reports, is 37.
In fact, 17 of the dream reports contain one or more of the ten 
semantically closest words.

\subsection{Name: Darryl, boyfriend}

\begin{verbatim}
train course bleachers dive hidden arthur easy supportive numbers tracks
 2.45  0.85    2.00    2.21  2.21   0.60  2.49    1.47     1.12    1.84
\end{verbatim}

The number of dream report texts with this name, from the 405 dream reports, is 21.
In fact, 8 of the dream reports contain one or more of the ten 
semantically closest words.

\subsection{Name: Lucy, friend (female)}

\begin{verbatim}
boyfriend song offers andrea leader ears rehearsal wrinkles counselors elizabeth
  3.10    3.13  2.94   2.61   1.54  2.22   2.96      2.68     2.78       3.11
\end{verbatim}

The number of dream report texts with this name, from the 405 dream reports, is 22.
In fact, 9 of the dream reports contain one or more of the ten 
semantically closest words.

\subsection{Name: Jake, brother}

\begin{verbatim}
shower age valerie parade shampoo  ex  fur program meaning curls
 2.75  2.83  2.02   2.65   1.60   2.39 1.83  2.35   2.84   2.85
\end{verbatim}

The number of dream report texts with this name, from the 405 dream reports, is 15.
In fact, 11 of the dream reports contain one or more of the ten 
semantically closest words.

\subsection{For name ``mother'', the succession of dream report texts}

The name ``mother'' is in 108 dream reports.  As noted above, just 24 of these dream 
reports contain one or more of the ten semantically closest words, in an overall sense. 
These words here, for this name, ``mother'', are: ``doors'', ``sleepy'', ``stood'', 
``nose'', ``nurse'', ``invisible'', ``clearly'', ``engagement'', ``king'', and ``mid''. 
These were determined as the closest words to ``mother'', from our word corpus other than the 11 
names, consisting of 1557 words.

The following lists the dream report, by its stated sequence number, followed
by the day, month and last two digits of the year. (Note that the first date here, is 
as stated on the original data. I.e., day and month are unspecified, and the year is 1977).   
Then the values, 0 and otherwise,
these are the number of occurrences, including non-occurrence = 0, of the word 
in the dream report.  

{\small 
\begin{verbatim}
Seq. Day-month-          The ten semantically closest words to name, "mother"
no.  year
                doors sleepy stood nose nurse invisible clearly engagement king mid
0052 ??_??_77     0      0     4    0     5         0       0          0    0   0
0107 29_09_80     0      1     0    0     0         0       0          0    0   0
0129 08_11_80     0      0     0    1     0         0       0          0    0   0
0243 19_01_81     0      1     0    0     0         0       0          0    0   0
0269 01_02_81     5      0     0    0     0         0       0          0    0   0
0288 14_02_81     0      3     0    0     0         0       1          0    0   0
0322 27_02_81     2      0     0    0     0         0       0          1    3   0
0339 07_03_81     0      0     0    0     0         1       0          0    0   2
0348 12_03_81     0      0     0    1     0         0       0          0    0   0
0370 23_03_81     0      1     0    0     0         0       0          0    0   0
0376 27_03_81     0      0     0    0     0         3       0          0    0   0
0412 17_04_81     1      0     0    0     0         0       0          0    0   0
0440 05_05_81     0      0     0    0     0         0       0          1    0   0
0442 06_05_81     0      0     0    1     0         0       0          0    0   0
0497 14_10_82     0      0     1    0     0         0       1          0    0   0
0507 01_11_82     0      0     0    1     0         0       0          0    0   0
0576 21_10_84     0      0     0    0     0         0       1          0    0   0
0715 15_03_85     0      0     0    0     0         0       0          0    1   0
0843 22_09_86     0      0     0    0     0         0       0          0    0   1
0920 17_05_88     0      0     0    0     0         0       1          0    0   0
0946 03_06_88     1      0     0    0     0         0       0          0    0   0
1020 21_07_88     0      1     0    0     0         0       0          2    0   0
1081 25_10_88     0      0     0    0     0         0       0          0    0   1
1262 16_02_89     0      0     0    2     0         0       0          0    0   0
\end{verbatim}
}

While there is not a great deal of presence of words here, nonetheless it may be
relevant in practice to avail of the relatively general expressions.  Some words
could become useful to seek trends with.  Here the word ``nurse'' was only used 
early in this context. Words such as ``doors'', ``sleepy'' recur.

All in all, this analysis is based on very general and overall context.  Hence 
the most general objective is to check out the balance of all that is relevant, 
to map out some of the most salient issues, and pursue general issues and themes.

\section{Study of Mother}
\label{sect6}

We take the 108 dream reports relating to the name ``mother'', that had, initially,
a word corpus of 1568 words.  The number of non-zero frequency of occurrence values
was 6.27\%.  Some of these words, collected from all of the dream reports under 
consideration here, had not got entries in dream reports relating to ``mother''. 
That reduced the number of words to 1433.  To have relevance, words that were at 
least occurring five times or more were determined and, so, the analysis is to be 
carried out on the 108 dream reports, relating to the word ``mother'', and with 
frequencies of occurrence for 662 words. 

The Correspondence Analysis, with the full dimensionality of the semantic, factor
space being 107, the cumulative percentages of inertia for the initial factors 
are: 3.27, 5.33, 7.31, 9.25, 11.13, 12.99,14.81, 16.59, 18.32, 20.05, and so on.

\begin{figure}
\includegraphics[width=12cm]{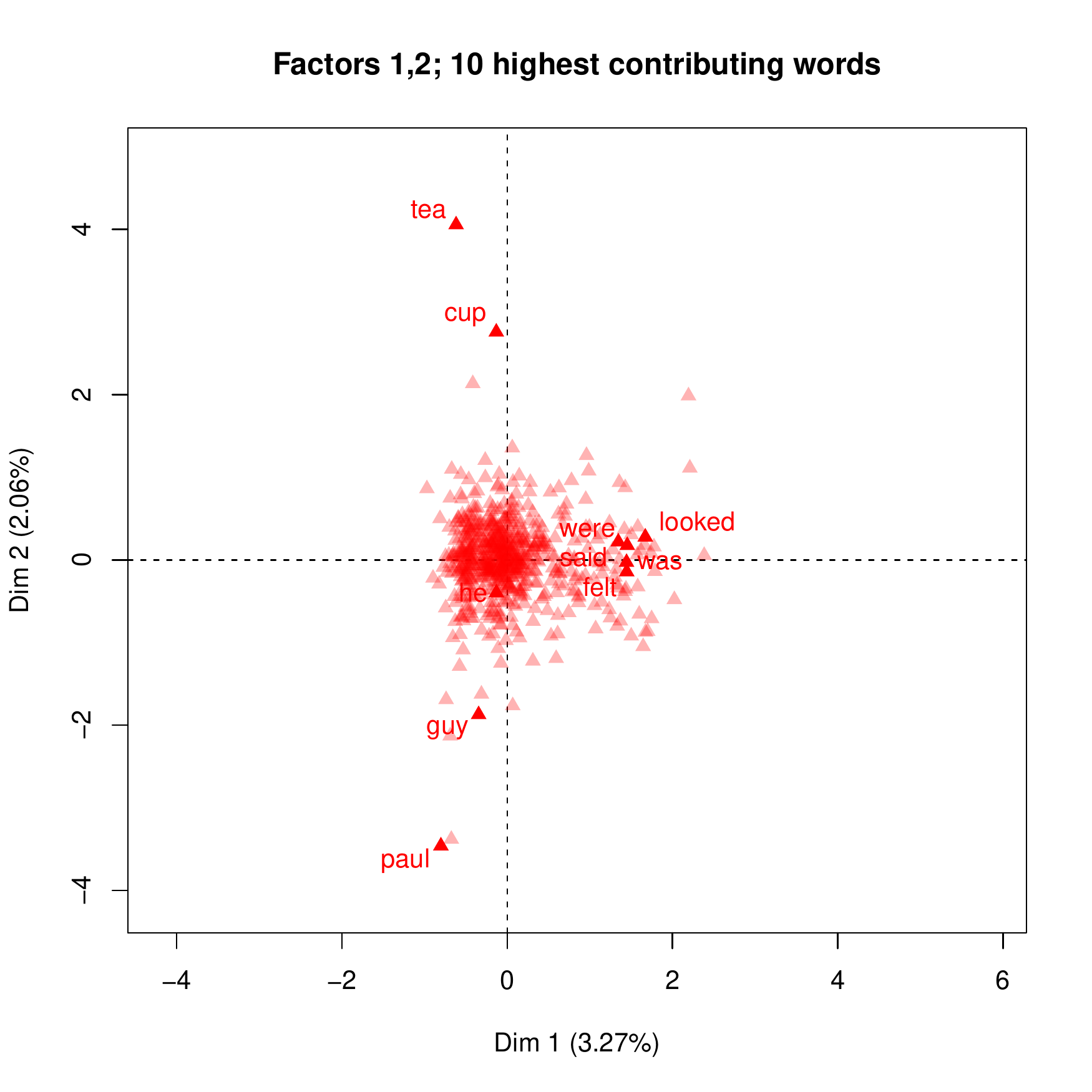}
\caption{The ten highest contributing words are displayed in the principal factor
plane.}
\label{fig7}
\end{figure}

In Figure \ref{fig7}, on the positive half axis of factor 1, there are these words:
``were'', ``looked'', ``said'', ``was'', ``felt''. Near the origin there it the word 
``he''.  For factor 2, on the positive half axis, there are the words ``tea'', ``cup'',
and on the negative half axis, there are the words ``paul'' and ``guy''. 

\begin{figure}
\includegraphics[width=12cm]{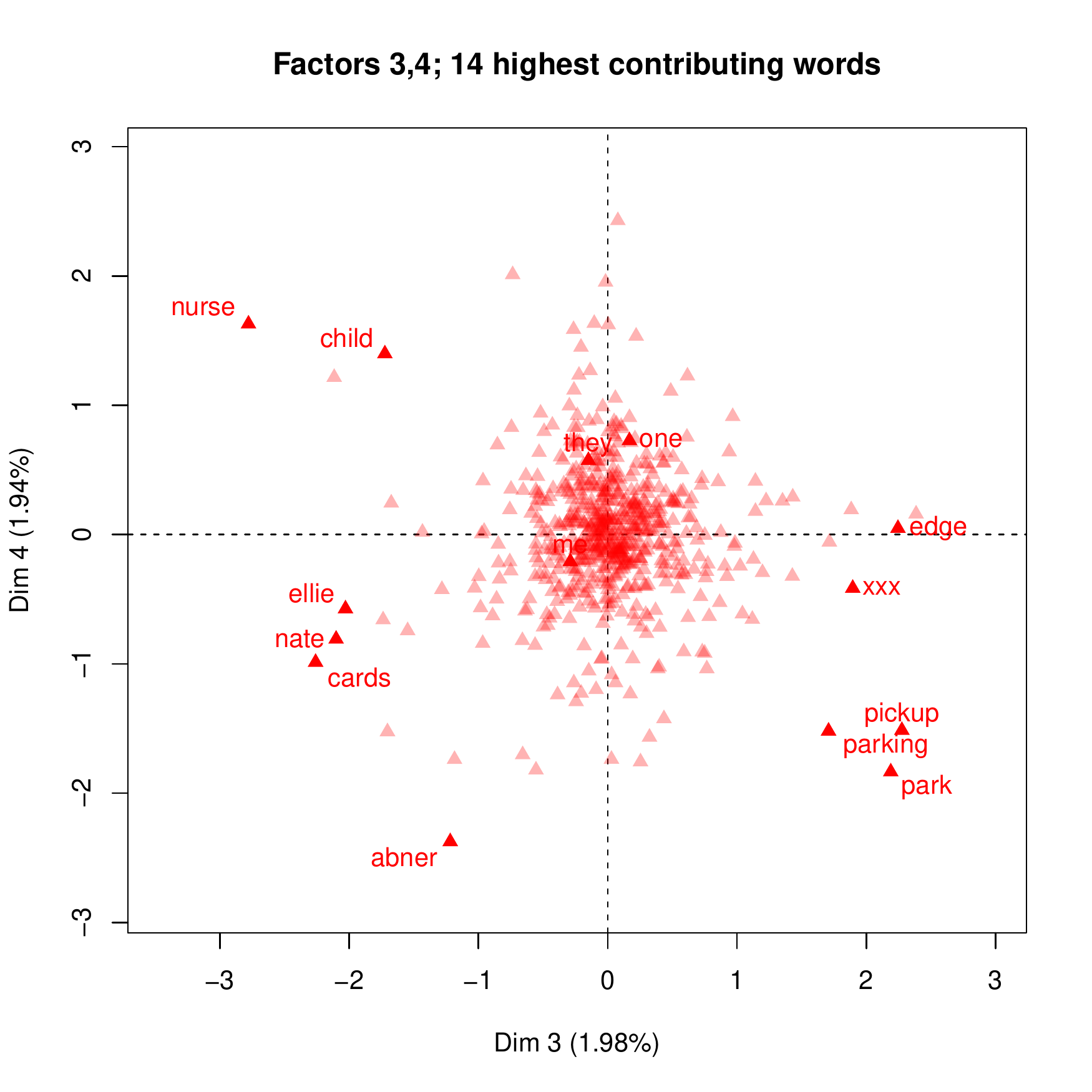}
\caption{The fourteen highest contributing words are displayed in the factor plane
of factors 3 and 4.}
\label{fig8}
\end{figure}

Figure \ref{fig8} displays the 14 highest contributing words for factors 3 and 4. 
Near the origin, on the positive side of factor 4, ``they'', ``one''; and on the 
negative side of factor 4, ``me''.  For factor 3, on the positive side, ``edge'',
``xxx'' (a word used in quite a few of the dream reports), ``pickup'', ``parking'', 
``park'', and the latter three are negative on factor 4.  The lower left quadrant 
has ``ellie'', ``nate'', ``cards'', ``abner''.  The upper right quadrant has 
``nurse'', ``child''.   Just for information about these words, all upper cases have
been put to lower case, in the analysis, and this expression is in this dream report 
(identifier number 0038, with the date, 23 September 1976), ``cousin Abner or Nate''.

\begin{figure}
\includegraphics[width=12cm]{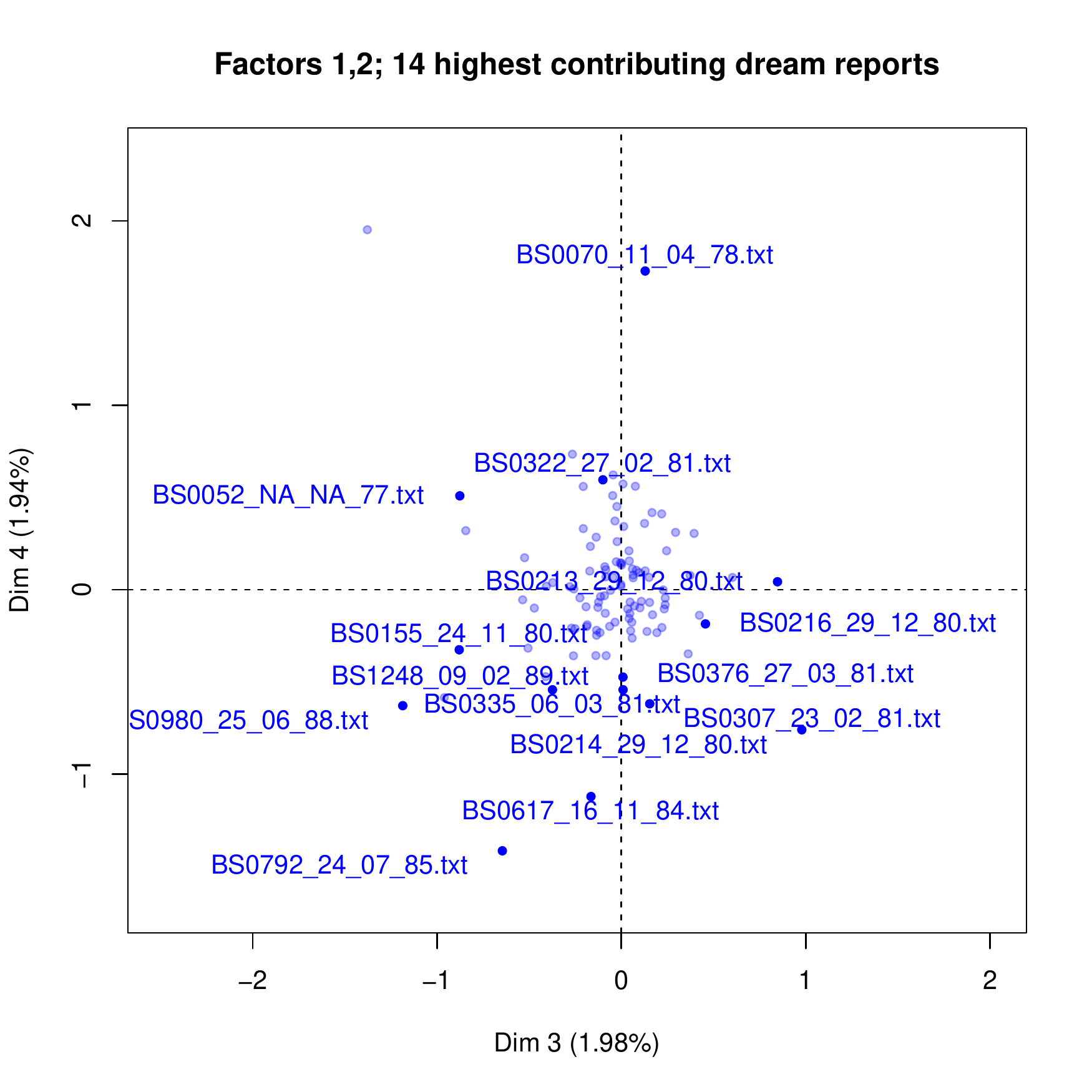}
\caption{The fourteen highest contributing dream reports are displayed in the 
principal factor plane.}
\label{fig9}
\end{figure}

Figure \ref{fig9} displays the principal factor plane with the highest contributing 
dream reports.  The labelling is the letter ``BS'', then the sequence number of the
dream report, followed then by its date, expressed as day, month and final two digits
of the year. 

\begin{figure}
\includegraphics[width=12cm]{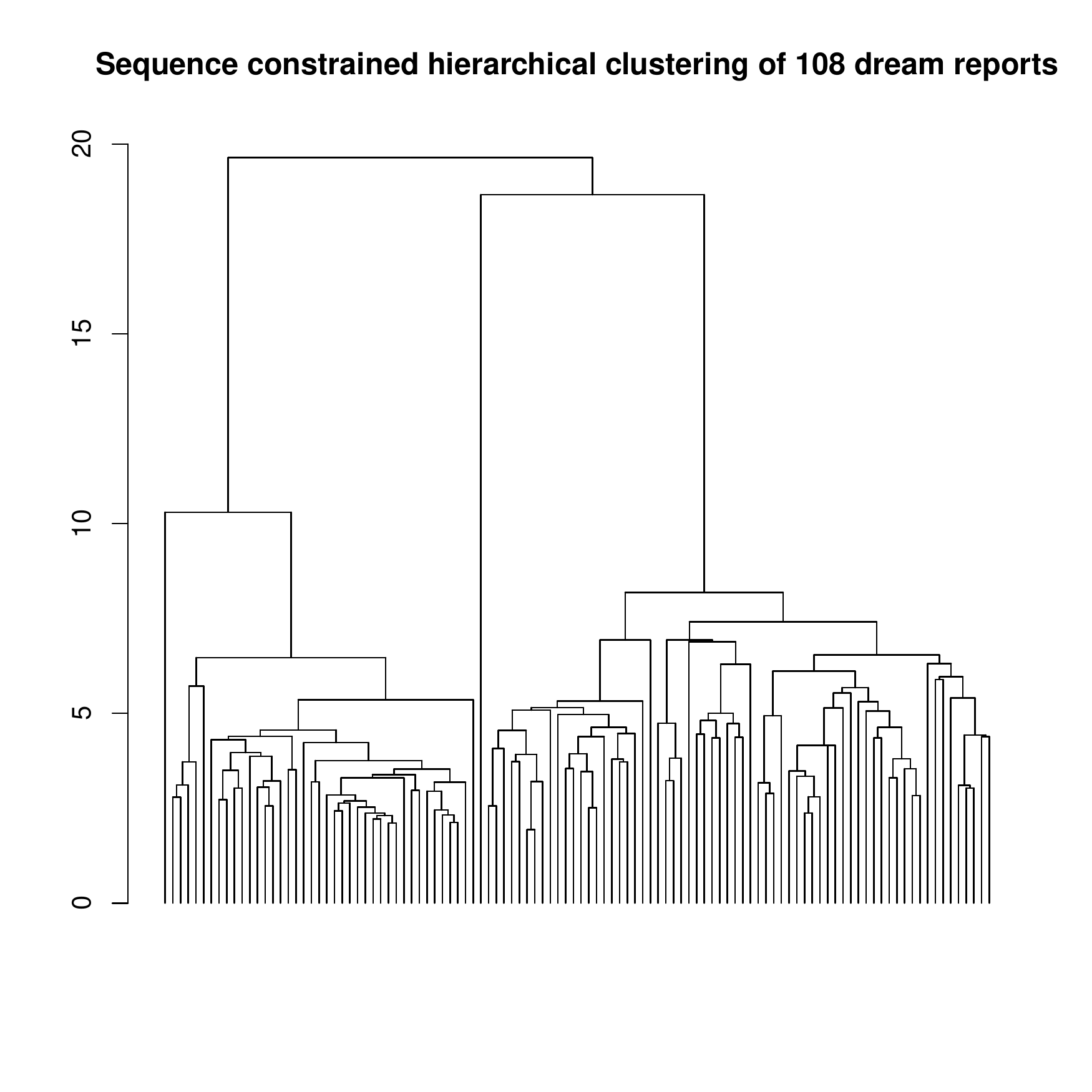}
\caption{Hierarchical clustering of the 108 dream reports, using the full dimensional
factor space, and having the chronological constraint.}
\label{fig10}
\end{figure}

Hierarchical clustering of the dream reports, subject to their sequencing, which 
is chronological, this is displayed in Figure \ref{fig10}.  A very interesting 
aspect of the dendrogram here is how different the first dream report is from the
the 2nd to the 39th.  Then comes a very major discontinuity, for the 42nd.  
The preceding and the following dream reports, relative to the 42nd are:
0392, 05\_04\_81;  0402, 11\_04\_81; 0408, 15\_04\_81.txt. 
The 40th dream report is very limited in its small number of words: 
``Somebody wore her mother's wedding gown.''  The previous dream report has 40 
word and the one after it has 282 words.  Sometimes successive dream reports are 
for the same date, but not here.  

\section{Conclusion}

An alternative approach to addressing the CCRT theme could certainly be 
questionnaire-based surveying.  This could well include both scale-based 
question responses and also free-text responses.  But better methodology,
along the lines of all that is in this paper, is to use narratives and 
accounting for a patient's or anyone's behavioural practices, and, quite
possibly also, for mental health.  

While very much that is involved in the data here is categorical, also 
termed qualitative data, then mapping into the factor space, that fully takes 
care of semantics, in effect is quantifying or quantification of our data.  

In this paper, there was demonstration of determining the most relevant words,
or derived terms, in the text source for analysis. The focus of the analysis, 
and studying balance and dominance of the data's contents, and very possibly, 
contextual description, all are at issue here.  Certain words or terms may be 
of importance, for example because they may express underpinning emotions
(hence related with responses of the type W, in agreement with CCRTs), but 
also the very general semantics, that are at issue in this article, these may
be very general in their application in a general context.  

Also it was noted how hierarchical clustering very well displays continuity or 
major change in the chronology of the narrative.  Once the input data has been 
well formatted for the analysis to be carried out on it, all that has been used
in this paper has been computational efficient and, we may note, all was 
implemented and used in the R software environment.

The final conclusion is to note the potentially great benefit, relevance and 
importance of such analytics fitting CCRT method with Geometric Data Analysis 
which have shown that, from clustering semantics coming from the analysis of the 
various factor spaces, REs are centred just around responses of the type W, as well 
as of the type RO and RS, if one takes into account the social group of persons 
considered and investigated in the text and drawn from the Barbara Sanders archive. 

\section*{References}

\begin{enumerate}


\item 
Aylward, P. (2012).  {\em Understanding Dunblane and Other Massacres: Forensic 
Studies of Homicide, Paedophilia, and Anorexia}, London: Karnac Books. 

\item
Domhoff, G. (2003). 
{\em The Scientific Study of Dreams: Neural Networks, Cognitive Development
and Content Analysis}. Washington D.C.: American Psychological Association.

\item
Domhoff, G. (2006). ``Barb Sanders: Our best case study
to date, and one that can be built upon''.  Retrieved from \\
http://www2.ucsc.edu/dreams/Findings/barb sanders.html.

\item Domhoff, G.W. \& Schneider, A. (2008). 
``Studying dream content using the archive and 
search engine on DreamBand.net'', {\em Consciousness and Cognition}, 17, 1238--1247. 
Retrieved from \\ 
https://www2.ucsc.edu/dreams/Library/domhoff\_2008c.html

\item Barbara Sanders Interview. ``Dreams and Waking Life: Interview Information 
to Accompany the Dream Journal of Barbara Sanders'', 53 pp. \\ 
Retrieved from \\
https://www2.ucsc.edu/dreams/Findings/barb\_sanders/barb\_sanders\_interviews.pdf

\item
DreamBank (2004). ``Repository of dream reports'', 2004, www.dreambank.net.


\item Laplanche, J. (1985). {\em Fantasme originaire, fantasmes des origines, origines du 
fantasme}, Paris: Hachette.

\item 
Laplanche, J. (1999). Interview.  Retrieved from \\  
https://www.radicalphilosophyarchive.com/  \\
wp-content/files\_mf/rp102\_interview\_laplanche.pdf

\item 
Luborsky, L. (1984). {\em Principles of Psychoanalytic Psychotherapy. A Manual for 
Supportive-Expressive Treatment}, New York: Basic Books Inc.  

\item
Luborsky L. \& Crits-Christoph, P. (1990).  {\em Understanding Transference, The CCRT 
Method}, New York: Basic Books Inc.  

\item 
Murtagh, F. (2014a). ``Pattern recognition in mental processes: determining vestiges of 
the subconscious through ultrametric component analysis'', in eds. S. Patel, Y. 
Wang, W. Kinsner, D. Patel, G. Fariello and L.A. Zadeh, {\em Proc. ICCI*CC 2014, 
2014 IEEE 13th International Conference on Cognitive Informatics and Cognitive 
Computing}, pp. 155-161. (Joint organiser and chair of session A5, Computational 
Psychoanalysis).

\item
Murtagh, F. (2014b). ``Pattern recognition of subconscious underpinnings of cognition 
using ultrametric topological mapping of thinking and memory'', {\em International 
Journal of Cognitive Informatics and Natural Intelligence} (IJCINI), 8(4), 1--16. 
(Also Guest Editor of this Special Issue on Computational 
Psychoanalysis.)

\item Murtagh, F. (2015). ``Correspondence Factor Analysis of Big Data Sets:
A Case Study of 30 Million Words; and Contrasting Analytics using Apache Solr and
Correspondence Analysis in R'', https://arxiv.org/abs/1507.01529

\item 
Murtagh, F. \& Iurato, G. (2016). ``Human behaviour, benign or malevolent: understanding 
the human psyche, performing therapy, based on affective mentalization and 
Matte-Blanco's bi-logic'', {\em Annals of Translational Medicine}, 4(24). \\
(Article in this issue: http://atm.amegroups.com/issue/view/507 ) 

\item Murtagh, F. and Iurato, G. (2017). ``Visualization of Jacques Lacan's registers 
of the psychoanalytic field, and discovery of metaphor and of metonymy. Analytical case 
study of Edgar Allan Poe's ``The Purloined Letter'', {\em Language and Psychoanalysis}, 
6(2), 26--55.

\end{enumerate} 

\end{document}